\documentclass[letterpaper]{article} 
\usepackage[preprint]{aaai}  
\usepackage[hyphens]{url}  
\usepackage{graphicx} 
\usepackage{natbib}  
\usepackage{caption} 
\usepackage{algorithm}
\usepackage{algorithmic}
\usepackage{amsmath}

\usepackage{multirow}
\usepackage{amssymb}

\usepackage{newfloat}
\usepackage{listings}
\DeclareCaptionStyle{ruled}{labelfont=normalfont,labelsep=colon,strut=off} 
\floatstyle{ruled}
\newfloat{listing}{tb}{lst}{}
\floatname{listing}{Listing}

\usepackage{booktabs}
\title{AgenticVAU: Multi-Agent Explore–Verify Reasoning for Video Anomaly Understanding}
\author{
    Yuxiang Duan\textsuperscript{\rm 1},
    Huining Li\textsuperscript{\rm 1},
    Ao Li\textsuperscript{\rm 2},
    Shuai Feng\textsuperscript{\rm 3},
    Lanju Kong\textsuperscript{\rm 1},\\
    Ning Liu\textsuperscript{\rm 1},
    Jian Zhang\textsuperscript{\rm 4},
    Xingdong Sheng\textsuperscript{\rm 4},
    Yuntao Du\textsuperscript{\rm 1}\corresponding
}
\affiliations{
    \textsuperscript{\rm 1}C-FAIR \& School of software, Shandong University
    \textsuperscript{\rm 2}Renmin University of China \\
    \textsuperscript{\rm 3}Nanjing Agricultural University
    \textsuperscript{\rm 4}Lenovo Research

}

\begin{document}

\maketitle

\begin{abstract}
Video anomaly understanding (VAU) focuses on comprehensively interpreting abnormal events in videos, requiring models to identify anomalous occurrences, discover their supporting evidence, and explain the underlying causes beyond simple anomaly detection.
Existing VAU methods often rely on specialized training or limited observations, restricting generalization or evidence coverage.
Although single-agent alternatives support adaptive video observation, they still integrate exploration, observation, and decision-making within a unified reasoning process, offering limited role specialization and structured evidence coordination.
To address these limitations, we present \textbf{AgenticVAU}, a training-free multi-agent framework that casts VAU as an explore--verify process, where the system first discovers potential anomalies and then verifies them through targeted observations.
To achieve this, four specialized agents are introduced to handle visual-rule construction, search planning, video observation, and final decision, respectively.
These agents communicate through an anchor registry, a shared evidence memory that binds each observation.
Guided by this agent framework, AgenticVAU interleaves broad temporal exploration, dense local verification, and cross-interval comparison until sufficient evidence is collected.
We conduct extensive experiments on the ECVA, UCF-Crime, and MSAD subsets of VAU-Bench, the results show that AgenticVAU outperforms zero-shot inference and reinforcement learning-based baselines, demonstrating the value of multi-agent collaboration for video anomaly understanding.
\end{abstract}

\section{Introduction}

Video anomaly understanding (VAU) aims to comprehensively interpret abnormal events in videos.
Beyond detecting or localizing an anomaly, a VAU system is expected to identify what happened, locate the relevant visual evidence, and explain why the event is abnormal.
Traditional video anomaly detection (VAD) methods primarily formulate anomaly analysis as frame- or segment-level prediction, producing anomaly scores or temporal boundaries to indicate where abnormal events occur~\citep{hasan2016learning,liu2018future,sultani2018realworld,tian2021rtfm}.
While effective for anomaly detection and localization, these outputs provide limited semantic information about the event itself.
Recent video-language models introduce language-based representations and generation into anomaly analysis, supporting semantic description, temporal grounding, and anomaly reasoning~\citep{du2024cuva,zhang2025holmesvau,zhu2025vaur1}.
This development shifts the focus from detecting abnormal patterns toward comprehensively interpreting anomalous events.

\begin{figure}[t]
\centering
\includegraphics[width=\linewidth]{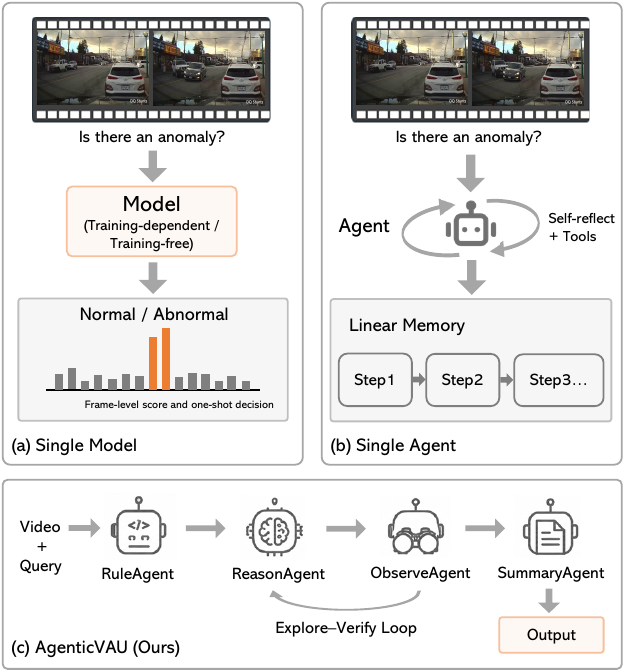}
\caption{
\textbf{Comparison of VAU paradigms.}
(a) Single-model methods make one-shot predictions from fixed observations.
(b) Single-agent methods rely on one agent to perform exploration, observation, evidence organization, and decision-making.
(c) AgenticVAU distributes these responsibilities among specialized agents within an explore--verify process.
}
\label{fig:motivation}
\vspace{-3mm}
\end{figure}

Despite this progress, adapting VAU methods to new scenarios and types of unseen anomalies remains challenging.
Many learning-based approaches rely on specialized anomaly data and task-specific optimization~\citep{zhang2025holmesvau,zhu2025vaur1}.
Training-free methods reduce this dependence, but generally infer from a limited set of sampled frames or derived visual descriptions~\citep{zanella2024lavad,yang2024anomalyruler}.
Agentic video systems provide a more flexible alternative by allowing the model to request additional observations during inference.
However, existing single-agent approaches unify exploration, observation, memory, and decision-making, limiting role specialization and structured coordination between evidence collection and anomaly judgment~\citep{yang2026panda}.
As illustrated in Figure~\ref{fig:motivation}, panel (a) shows single-model methods that make one-shot predictions from fixed observations, while panel (b) shows single-agent methods that rely on one agent to perform exploration, observation, evidence organization, and decision-making.
This motivates a training-free multi-agent framework that separates complementary responsibilities while maintaining a shared and structured evidence state, as illustrated in Figure~\ref{fig:motivation}(c).

To this end, we present \textbf{AgenticVAU}, a training-free multi-agent framework that casts VAU as an explore--verify process.
The system first explores the video to identify potential anomalous events and then evaluates them through targeted observations.
Four specialized agents are introduced to handle visual-rule construction, search planning, video observation, and final decision, respectively.
These agents communicate through an anchor registry, a shared evidence memory that records each observation together with its temporal scope, associated visual rule, and verification state.
Rather than treating all observations as an accumulated content history, the registry organizes them according to whether they support, oppose, or remain inconclusive about a candidate anomaly event.
This representation provides the agents with a compact evidence state for planning subsequent observations and forming the final prediction.

Guided by this agent framework, AgenticVAU can broaden its temporal search, revisit selected intervals with denser observations, or compare evidence from separated moments until sufficient evidence is collected.
We evaluate AgenticVAU on the ECVA, UCF-Crime, and MSAD subsets of VAU-Bench~\citep{zhu2025vaur1}.
The results show improvements over the evaluated direct zero-shot inference and reinforcement-fine-tuned baselines, with particularly large gains in binary and multi-class anomaly classification.
These results demonstrate the effectiveness of multi-agent collaboration in coordinating video exploration, evidence verification, and final decision-making for video anomaly understanding.

Our contributions are summarized as follows:
\begin{itemize}
\item We formulate VAU as a training-free multi-agent explore--verify process, in which video observations are adaptively collected according to the evidence required to verify candidate anomaly events.
\item We introduce a novel multi-agent system along with the anchor registry, a shared evidence memory that associates observations with visual rules and verification states, helping the system distinguish opposing evidence from observations that remain inconclusive.
\item Experiments on VAU-Bench show that AgenticVAU improves over the evaluated zero-shot and reinforcement-fine-tuned baselines across multiple VAU tasks, particularly on anomaly classification.
\end{itemize}

\section{Related Work}
\subsection{Video Anomaly Detection and Understanding}
Traditional video anomaly detection (VAD) methods mainly identify deviations from normal visual patterns.
Unsupervised approaches learn normality through reconstruction, future-frame prediction, or memory-based modeling~\citep{hasan2016learning,liu2018future,gong2019memorizing}, while weakly supervised methods localize anomalous segments from video-level supervision~\citep{sultani2018realworld,tian2021rtfm,chen2023mgfn,zhou2023urdmu,pu2024pel4vad,yang2024tpwng}.
Vision-language models further introduce language-aligned semantic representations into weakly supervised VAD~\citep{wu2024vadclip}, and open-vocabulary or open-world methods extend anomaly recognition to previously unseen event categories~\citep{wu2024ovvad,liu2025lagovad}.
These approaches have substantially advanced anomaly recognition and localization, although their outputs are commonly represented as anomaly scores, temporal boundaries, or event categories.

Recent studies extend VAD toward video anomaly understanding with LLMs and MLLMs.
LAVAD~\citep{zanella2024lavad} estimates anomaly scores from frame descriptions without task-specific training, whereas AnomalyRuler~\citep{yang2024anomalyruler} induces scene-specific rules from a small set of normal references.
HAWK~\citep{tang2024hawk} and CUVA~\citep{du2024cuva} study open-world and causation-oriented anomaly understanding, respectively, while Holmes-VAU~\citep{zhang2025holmesvau} considers multi-granular understanding of long videos.
VAU-R1~\citep{zhu2025vaur1} improves multiple VAU tasks through reinforcement fine-tuning, and FineVAU~\citep{pereira2026finevau} focuses on fine-grained, human-aligned evaluation.
More recently, PANDA~\citep{yang2026panda} introduces planning, tool use, reflection, and memory into generalist anomaly detection, while QVAD~\citep{bekit2026qvad} iteratively refines visual questions to guide training-free video inspection.
Anom-\(\pi\)~\citep{mo2026learning} further learns an active observation policy with temporal expansion, backtracking, and fine-grained sampling.
In contrast to these single-agent or learned-policy formulations, AgenticVAU separates rule construction, search planning, observation, and decision-making, and organizes the collected observations according to their evidential relation to candidate anomaly events.

\subsection{Agentic Video Reasoning}
Agentic video reasoning allows a language model to acquire additional video information during inference.
Following reasoning--action paradigms developed for language agents~\citep{yao2022react,shinn2023reflexion}, VideoAgent-style systems combine iterative frame retrieval with memory construction for long-video understanding~\citep{fan2024videoagent,wang2024videoagent}.
Subsequent methods incorporate uncertainty-aware observation and planning~\citep{zhi2025videoagent2}, adaptive or tool-guided video exploration~\citep{yang2025vca,zhang2025dvd}, and hierarchical multimodal memory~\citep{yin2026videoarm}.
LensWalk~\citep{li2026lenswalk} and VideoSeek~\citep{lin2026videoseek} further allow the observation scope or temporal granularity to be adjusted according to the current query and reasoning state.
These methods improve adaptive access to relevant video content, with memory commonly used to retain observations, summaries, tool outputs, uncertainty estimates, or multi-level video clues.
AgenticVAU builds on active video observation while additionally associating each observation with its role in verifying or rejecting a candidate anomaly event, or leaving it unresolved.

Multi-agent methods distribute perception and reasoning among agents with different responsibilities.
VideoMultiAgents~\citep{kugo2025videomultiagents}, LVAgent~\citep{chen2025lvagent}, and LongVideoAgent~\cite{liu2026longvideoagent} coordinate multiple agents for video question answering and long-video reasoning.
MUPA~\citep{dang2025mupa} explores multiple grounding and reasoning paths, while SVAgent~\citep{yang2026svagent} and MACF~\citep{chen2026macf} investigate cross-modal collaboration and communication among specialized agents.
These studies indicate that role specialization can facilitate collection and integration of video information. Thus, AgenticVAU applies this principle to VAU by separating visual-rule construction, search planning, video observation, and final decision, along with a shared evidence state that guides subsequent exploration and evidence aggregation.

\section{Method}

\subsection{Overview}
Given a video \(V\) and a query \(Q\), video anomaly understanding requires the system to identify anomalous events, locate relevant visual evidence, and produce a task-specific output \(\hat{Y}\), such as an answer option, anomaly category, temporal interval, or textual explanation.
AgenticVAU addresses this process with four specialized agents: \emph{RuleAgent} constructs supporting and opposing visual rules, \emph{ReasonAgent} plans each search step, \emph{ObserveAgent} inspects the requested video content, and \emph{SummaryAgent} aggregates the collected evidence into the final output.
The agents communicate through an anchor registry that records each observation with its temporal support, associated visual rule, verification state, and evidence summary.

\begin{figure*}[ht]
\centering
\includegraphics[width=\textwidth]{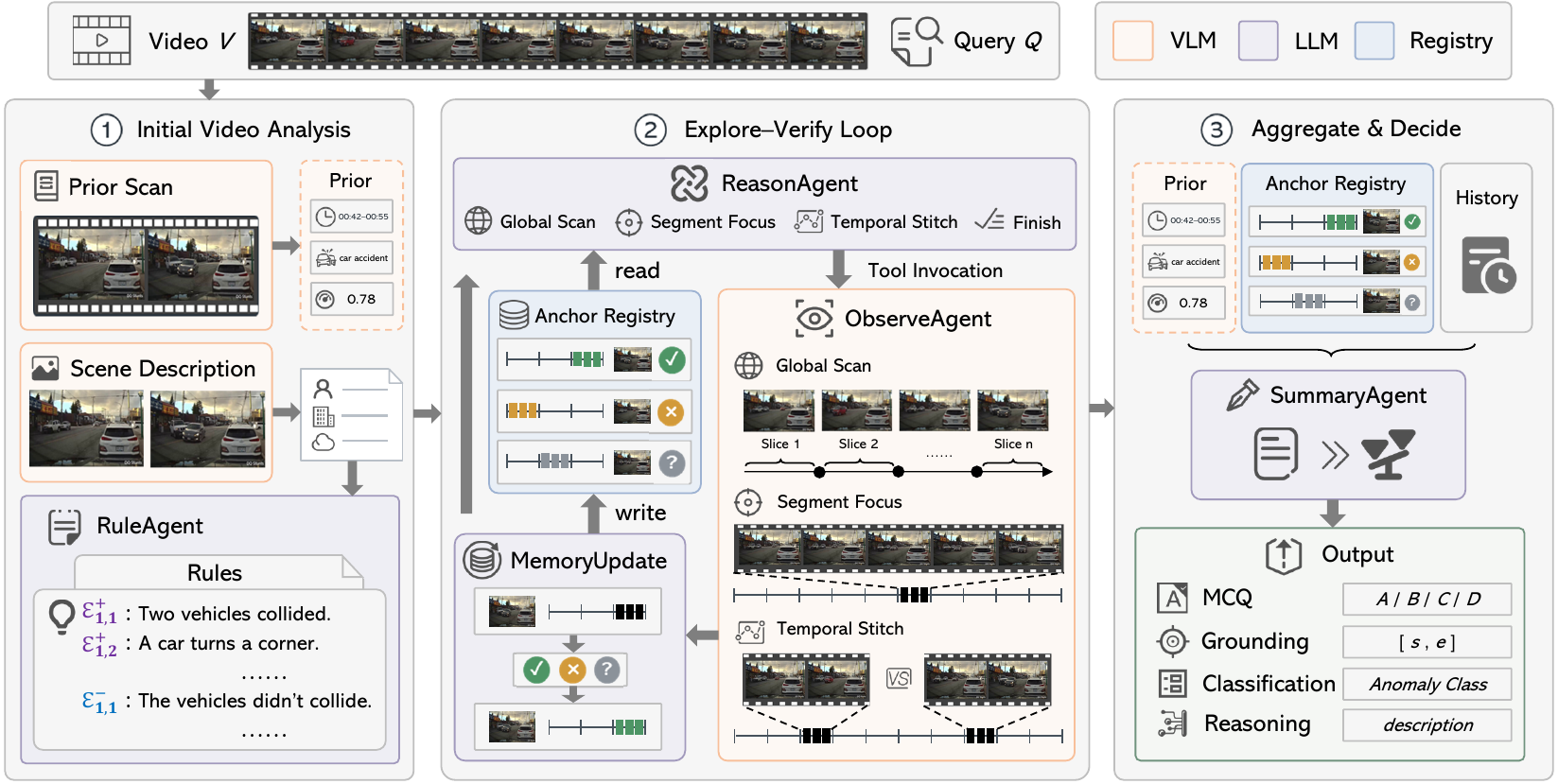}
\caption{
\textbf{Overview of AgenticVAU.}
Given a video \(V\) and a query \(Q\), the framework first obtains a video prior and constructs scene-aware visual rules, where
\(\mathcal{E}^{+}_{j,k}\) and \(\mathcal{E}^{-}_{j,k}\) denote the supporting and opposing visual criteria of rule \(r_j\).
During the Explore--Verify Loop, ReasonAgent plans each observation, ObserveAgent inspects the requested video content, and the resulting evidence is organized in the anchor registry.
SummaryAgent then aggregates the final evidence state into a task-specific prediction.
}
\label{fig:method}
\end{figure*}

As illustrated in Figure~\ref{fig:method}, AgenticVAU contains three stages.
During \emph{Initial Video Analysis}, the system performs a prior scan and generates a scene description, from which RuleAgent constructs scene-aware visual rules.
During the \emph{Explore--Verify Loop}, ReasonAgent determines what evidence is needed, ObserveAgent inspects the requested video content, and the MemoryUpdate records the resulting evidence.
Finally, during \emph{Aggregate and Decide}, SummaryAgent converts the final evidence state into the output.

\subsection{Initial Video Analysis}
Before iterative exploration, the system estimates where the queried event may occur and determines what visual evidence should be examined.
The initial analysis therefore provides two complementary forms of guidance: a query-conditioned prior for initializing temporal exploration and scene-aware visual rules for evaluating the events encountered during the search.

\paragraph{Prior Scan.}
The system first performs a query-conditioned scan from the full video.
The resulting prior \(P=F_{\mathrm{prior}}(V,Q)\) provides an estimate of the queried event, including an approximate temporal interval and a short event description.
Depending on the query, it may also contain a tentative anomaly category or answer option.

The prior allows exploration to begin from a potentially relevant interval rather than searching the entire video without guidance.
For example, Figure~\ref{fig:method} shows a prior interval of 00:42--00:55 with a tentative car-accident label.
Because this estimate is obtained from sparse observations and may be incorrect, \(P\) is used only to initialize exploration.
It is not treated as verified evidence or as the final prediction.

\paragraph{Scene Description.}
The interpretation of an event depends on its surrounding scene and typical activities.
The system therefore generates a query-agnostic scene description \(C=F_{\mathrm{scene}}(V)\), which summarizes the environment, relevant entities and objects, and common activities observed in the video.
It provides context for determining which visual patterns should support or oppose the event.
For example, an abrupt stop may indicate a collision at an intersection but be routine at a toll gate.
Without scene context, the resulting visual criteria may be too generic for reliable verification.

\paragraph{Visual Rule Generation.}
RuleAgent converts the query and scene description into a set of scene-aware visual rules:

\begin{equation}
\begin{aligned}
\mathcal{R}
&=
F_{\mathrm{rule}}(Q,C)
=
\{r_j\}_{j=1}^{N_r}, \\
r_j
&=
(\mathcal{E}^{+}_j,\mathcal{E}^{-}_j), \\
\mathcal{E}^{\pm}_j
&=
\{\mathcal{E}^{\pm}_{j,k}\}_{k=1}^{N^{\pm}_j}.
\end{aligned}
\label{eq:visual_rules}
\end{equation}

Here, \(N_r\) denotes the number of generated visual rules.
\(\mathcal{E}^{+}_j\) and \(\mathcal{E}^{-}_j\) are the sets of supporting and opposing visual criteria for rule \(r_j\), respectively, while \(N_j^{+}\) and \(N_j^{-}\) denote their corresponding numbers of criteria.
The term \(\mathcal{E}^{\pm}_{j,k}\) denotes the \(k\)-th criterion in the corresponding set.
Each criterion is expressed as a concrete visual condition that ObserveAgent can evaluate from selected frames.

For the collision example in Figure~\ref{fig:method}, a supporting criterion may be that two vehicles make physical contact, while an opposing criterion may be that they pass without contact.
Both sets are necessary because the absence of supporting evidence is not itself opposing evidence.
An occluded interaction leaves the rule unresolved, whereas clearly observing the vehicles pass without contact directly contradicts it.

\subsection{Explore--Verify Loop}

The initial analysis provides a search prior and visual criteria, but it does not establish whether the queried event is supported by sufficient evidence.
Moreover, different anomalies may require different temporal ranges, sampling densities, or comparisons between separated moments.
The Explore--Verify Loop therefore iteratively plans observations, inspects selected video content, and updates the shared evidence state.

Specifically, ReasonAgent identifies an unresolved evidence requirement at each step.
ObserveAgent then performs the requested inspection, and the registry updater incorporates the resulting observation into the anchor registry.
The loop then proceeds to the next iteration, forming a closed evidence-driven cycle.

\paragraph{Search Planning.}
The loop proceeds over a sequence of iterations indexed by \(t \in \{1,2,\ldots,T\}\), where each iteration collects one new video observation.
At the beginning of iteration \(t\), the anchor registry \(\mathcal{M}_{t-1}\) contains the structured evidence collected so far, while \(\mathcal{H}_{t-1}\) records the previous observation actions and their corresponding results.
Based on the current evidence state, ReasonAgent either terminates the loop with the \emph{Finish} action or generates the next observation action:

\begin{equation}
\begin{aligned}
a_t
&=
\pi_{\mathrm{reason}}
\left(
Q,
P,
\mathcal{R},
\mathcal{M}_{t-1},
\mathcal{H}_{t-1}
\right) \\
&=
\left(
r_t,
q_t,
J_t,
g_t,
s_t
\right).
\end{aligned}
\label{eq:search_action}
\end{equation}

Here, \(a_t\) is the observation instruction at iteration \(t\), specifying what evidence should be examined, where it should be sought, and how it should be collected.
In particular, \(r_t\in\mathcal{R}\) is the visual rule being examined, and \(q_t\) is the corresponding visual question posed to ObserveAgent.
The temporal scope \(J_t\) specifies intervals to inspect.
The variable \(g_t\) selects the observation tool, while \(s_t\) specifies its sampling configuration.
When no reliable candidate interval has been identified, ReasonAgent expands temporal coverage.
When a candidate interval lacks sufficient local evidence, it revisits that interval with denser sampling.
When verification depends on relations between separated moments, it requests a cross-interval comparison.

\paragraph{Multi-Granularity Video Observation.}
The three observation tools expose different types of temporal evidence: Global Scan provides broad coverage, Segment Focus provides local detail, and Temporal Stitch reveals relations between separated moments. After executing the selected tool, ObserveAgent returns a rule-specific observation \(o_t\) that describes the relevant visual evidence and its temporal support.

\textbf{Global Scan} divides a long temporal range into consecutive short slices and samples frames from each slice.
ObserveAgent examines these samples in temporal order to identify which parts of the range contain visual cues related to the queried event.
It is used to locate candidate intervals for further inspection rather than to verify the event directly.

\textbf{Segment Focus} samples a selected continuous interval at a higher temporal density.
The denser observation reveals short actions and state changes, allowing ObserveAgent to check whether the supporting or opposing criteria of the active rule are visible.
Segment Focus is therefore used to verify or reject a candidate event and, when possible, refine its temporal boundaries.

\textbf{Temporal Stitch} samples frames from multiple disjoint intervals and places them within the same observation context in temporal order.
This allows ObserveAgent to directly compare separated moments when the active rule depends on their relation, such as a before--after change, a repeated action, or the consequence of an earlier event.
Temporal Stitch is used when the required evidence cannot be established from a single continuous interval.

\paragraph{Anchor Registry Update.}
Different observations may examine the same event at different temporal ranges or sampling densities.
To organize these observations, AgenticVAU represents each candidate event as an anchor and stores its current evidence in the anchor registry:

\begin{equation}
\begin{aligned}
o_t
&=
F_{\mathrm{observe}}(V,a_t), \\
A_t
&=
F_{\mathrm{anchor}}(o_t,a_t)
=
\left(
\mathcal{I}_t,
r_t,
z_t,
e_t
\right), \\
\mathcal{M}_t
&=
U_{\mathrm{mem}}
\left(
\mathcal{M}_{t-1},
A_t
\right).
\end{aligned}
\label{eq:registry_update}
\end{equation}

Here, \(\mathcal{I}_t\) is the temporal interval or intervals covered by the evidence, \(r_t\) is the visual rule being evaluated, \(e_t\) summarizes the observed evidence, and \(z_t\) records its verification state.
The registry is initialized as \(\mathcal{M}_0=\varnothing\).

An anchor is marked as \emph{candidate} when Global Scan identifies a suspicious interval that still requires verification.
It becomes \emph{verified} when the supporting criteria of \(r_t\) are observed, or \emph{rejected} when its opposing criteria are observed.
If the inspected content is insufficient or ambiguous, the anchor is marked as \emph{unclear}.

For the collision rule, observing that two vehicles collide verifies the anchor, whereas observing that they do not collide rejects it.
If the available frames do not support either conclusion, the anchor remains unclear.
This distinction prevents insufficient evidence from being treated as evidence against the event.

When a new observation refers to the same rule and overlaps an existing anchor in time, \(U_{\mathrm{mem}}\) merges the new evidence into that anchor.
The updated evidence may change a candidate or unclear anchor to \emph{verified} or \emph{rejected}; otherwise, it remains unresolved for further inspection.

\paragraph{Loop Control.}
After each registry update, ReasonAgent decides whether more evidence is needed.
The loop terminates when ReasonAgent selects the \emph{Finish} action, the maximum number of steps is reached, or several consecutive observations produce no useful registry update.
For multiple-choice questions, it may also stop once the collected evidence is sufficient to distinguish among the answer options.

\subsection{Aggregate and Decide}

The Explore--Verify Loop ends with a collection of observations and a structured anchor registry rather than a task-formatted answer.
SummaryAgent therefore converts the accumulated evidence into the final answer.

\paragraph{Evidence Aggregation.}
At the final iteration \(T\), SummaryAgent receives the query, initial prior, final anchor registry, and complete interaction history:

\begin{equation}
\hat{Y}
=
F_{\mathrm{summary}}
\left(
Q,
P,
\mathcal{M}_T,
\mathcal{H}_T
\right),
\label{eq:final_prediction}
\end{equation}

where \(\mathcal{M}_T\) contains the structured evidence collected during exploration, and \(\mathcal{H}_T\) records the sequence of observation actions and their results.
The query \(Q\) specifies the required task and output format, while the prior \(P\) provides only the initial context for the search.

SummaryAgent reads each anchor with its associated visual rule, temporal support, verification state, and evidence summary.
A \emph{verified} anchor indicates that the supporting criteria of the corresponding rule have been observed, whereas a \emph{rejected} anchor indicates that its opposing criteria have been observed.
Candidate and unclear anchors are retained as unresolved evidence and are not treated as either support for or evidence against the queried event.
SummaryAgent then combines these anchors across their temporal intervals to form the final decision.
The prior \(P\) provides only the initial search context; if it conflicts with the evidence collected during the loop, the resolved anchors in \(\mathcal{M}_T\) take precedence.

\paragraph{Task Output.}
Based on the aggregated evidence, SummaryAgent returns an answer option for multiple-choice question answering, a start and end time for temporal grounding, an anomaly category for classification, or an evidence-based explanation for anomaly reasoning.
All requested outputs are derived from the same set of task-relevant anchors to maintain consistency across tasks.

\section{Experiment}
\subsection{Experimental Setup}

\paragraph{Datasets and Tasks.}
We conducted experiments on VAU-Bench~\citep{zhu2025vaur1}, which consists of three video anomaly datasets: ECVA~\citep{du2024ecva}, UCF-Crime~\citep{sultani2018realworld}, and MSAD~\citep{zhu2024msad}.
Following the official evaluation protocol, we report results separately for each dataset for four tasks: multiple-choice question answering, temporal anomaly grounding, anomaly reasoning, and anomaly classification.

\paragraph{Evaluation Metrics.}
For multiple-choice question answering, we report the answer accuracy.
Temporal grounding is evaluated by mean temporal intersection over union (mIoU) and recall at IoU thresholds of \(0.3\), \(0.5\), and \(0.7\).
For anomaly reasoning, we adopt VAU-Eval~\citep{zhu2025vaur1}, which measures classification correctness (CLS), key-concept matching (KM), fluency (FLU), informativeness (INF) and factual consistency (FAC).
Each VAU-Eval dimension is scored on a scale from 0 to 10, resulting in a maximum total score of 50.
For anomaly classification, we report binary accuracy and multi-class accuracy.

\paragraph{Implementation Details.}
We use Qwen2.5-VL-3B~\citep{qwen2.5vl} as the video observation model and DeepSeek-V4-Pro~\citep{xu2026deepseekv4} as the reasoning model for rule generation, search planning, memory update, and final prediction.
The maximum number of Explore--Verify steps is set to 15.
Unless otherwise specified, the same inference configuration is used across all three datasets.
All experiments were conducted on NVIDIA A800 GPUs with 80 GB memory.

\paragraph{Compared Methods.}
We compare AgenticVAU with Qwen2.5-VL-3B and VAU-R1~\citep{zhu2025vaur1}.
Qwen2.5-VL-3B performs zero-shot inference, while VAU-R1 applies reinforcement fine-tuning to the same backbone for video anomaly understanding.
AgenticVAU does not require task-specific training and instead performs iterative evidence exploration and verification at inference time.

\begin{table}[t]
    \centering
    \small
    \setlength{\tabcolsep}{1pt}
    \renewcommand{\arraystretch}{1}
    \begin{tabular*}{\linewidth}{
        @{\extracolsep{\fill}}llccc@{}
    }
        \toprule
        Dataset & Model & MCQ Acc. & Bin. Acc. & Multi. Acc. \\
        \midrule

        \multirow{3}{*}{ECVA}
        & Qwen2.5-VL-3B & 85.58 & \underline{52.83} & 30.16 \\
        & VAU-R1         & \underline{89.53} & 49.66 & \underline{30.61} \\
        & AgenticVAU     & \textbf{91.10} & \textbf{76.94} & \textbf{48.86} \\

        \midrule

        \multirow{3}{*}{UCF-Crime}
        & Qwen2.5-VL-3B & 91.63 & \underline{64.54} & \underline{58.57} \\
        & VAU-R1         & \underline{92.03} & 62.55 & 57.77 \\
        & AgenticVAU     & \textbf{94.42} & \textbf{94.42} & \textbf{85.66} \\

        \midrule

        \multirow{3}{*}{MSAD}
        & Qwen2.5-VL-3B & 85.83 & 79.17 & 69.58 \\
        & VAU-R1         & \underline{88.33} & \underline{82.08}
                         & \underline{71.25} \\
        & AgenticVAU     & \textbf{91.25} & \textbf{90.00}
                         & \textbf{75.42} \\

        \bottomrule
    \end{tabular*}
    \caption{
        Comparison on multiple-choice question answering and anomaly classification. Bin. Acc. denotes normal--abnormal classification accuracy, while Multi. Acc. denotes multi-class anomaly classification accuracy. The best result is shown in bold and the second best is underlined.
    }
    \label{tab:main_semantic}
\end{table}

\begin{table}[t]
    \centering
    \small
    \setlength{\tabcolsep}{0.5pt}
    \renewcommand{\arraystretch}{1}
    \begin{tabular*}{\linewidth}{
        @{\extracolsep{\fill}}llcccccc@{}
    }
        \toprule
        Dataset & Model & CLS & KM & FLU & INF & FAC & Total \\
        \midrule

        \multirow{3}{*}{ECVA}
        & Qwen2.5-VL-3B
        & \underline{2.21} & \underline{2.58}
        & \underline{8.33} & \underline{5.02}
        & \underline{2.75} & \underline{20.89} \\
        & VAU-R1
        & 1.45 & 2.24 & 8.05 & 4.32 & 2.39 & 18.45 \\
        & AgenticVAU
        & \textbf{7.38} & \textbf{6.03}
        & \textbf{8.72} & \textbf{5.45}
        & \textbf{5.23} & \textbf{32.81} \\

        \midrule

        \multirow{3}{*}{\shortstack{UCF-\\Crime}}
        & Qwen2.5-VL-3B
        & 4.31 & 2.88 & \underline{8.70}
        & 5.95 & 3.27 & 25.10 \\
        & VAU-R1
        & \underline{4.42} & \underline{2.98}
        & \textbf{8.71} & \underline{5.98}
        & \underline{3.39} & \underline{25.49} \\
        & AgenticVAU
        & \textbf{8.40} & \textbf{6.44}
        & 7.94 & \textbf{6.02}
        & \textbf{6.70} & \textbf{35.50} \\

        \midrule

        \multirow{3}{*}{MSAD}
        & Qwen2.5-VL-3B
        & 5.77 & 5.24 & 9.02 & 6.74 & 5.70 & 32.47 \\
        & VAU-R1
        & \underline{5.97} & \underline{5.49}
        & \underline{9.05} & \textbf{6.84}
        & \underline{6.03} & \underline{33.38} \\
        & AgenticVAU
        & \textbf{7.07} & \textbf{6.80}
        & \textbf{9.18} & \underline{6.75}
        & \textbf{7.12} & \textbf{36.92} \\

        \bottomrule
    \end{tabular*}
    \caption{
        Comparison on anomaly reasoning using VAU-Eval. CLS, KM, FLU, INF, and FAC denote classification correctness, key-concept matching, fluency, informativeness, and factual consistency. Total is the sum of the five dimensions.
    }
    \label{tab:main_reasoning}
\end{table}

\subsection{Main Results}
\paragraph{Multiple-Choice QA and Anomaly Classification.}
As shown in Table~\ref{tab:main_semantic}, AgenticVAU consistently achieves the best performance across all datasets and evaluation metrics.
The gains on multiple-choice questions are steady, while substantially larger improvements are observed for binary and multi-class classification, particularly on ECVA and UCF-Crime.
For example, AgenticVAU improves binary accuracy by 24.11\% and 29.88\% on ECVA and UCF-Crime, respectively, over the strongest baseline.

\paragraph{Anomaly Reasoning.}
Table~\ref{tab:main_reasoning} shows that AgenticVAU achieves the best overall VAU-Eval scores across all three datasets.
The improvements mainly arise from classification correctness, key-concept matching, and factual consistency, while fluency and informativeness remain generally comparable to the baselines.
This suggests that the collected observations not only support more accurate predictions, but also provide more relevant and visually grounded evidence for explaining anomalous events.

\begin{table}[t]
    \centering
    \small
    \setlength{\tabcolsep}{1pt}
    \renewcommand{\arraystretch}{1}
    \begin{tabular*}{\linewidth}{
        @{\extracolsep{\fill}}llcccc@{}
    }
        \toprule
        Dataset & Model & mIoU & R@0.3 & R@0.5 & R@0.7 \\
        \midrule

        \multirow{2}{*}{ECVA}
        & Qwen2.5-VL-3B
        & 14.21 & 17.16 & 6.47 & 3.23 \\
        & AgenticVAU
        & \textbf{25.48} & \textbf{31.25}
        & \textbf{23.25} & \textbf{15.25} \\

        \midrule

        \multirow{2}{*}{\shortstack{UCF-\\Crime}}
        & Qwen2.5-VL-3B
        & 10.91 & 15.32 & 6.45 & 3.23 \\
        & AgenticVAU
        & \textbf{15.42} & \textbf{19.97}
        & \textbf{7.03} & \textbf{4.21} \\

        \midrule

        \multirow{2}{*}{MSAD}
        & Qwen2.5-VL-3B
        & 21.27 & 30.00 & 10.83 & 4.17 \\
        & AgenticVAU
        & \textbf{30.80} & \textbf{35.00}
        & \textbf{26.67} & \textbf{20.00} \\

        \bottomrule
    \end{tabular*}
    \caption{
        Comparison on temporal anomaly grounding. We report mean
        temporal IoU (mIoU) and recall at IoU thresholds of \(0.3\),
        \(0.5\), and \(0.7\).
    }
    \label{tab:main_grounding}
\end{table}

\begin{table}[t]
    \centering
    \small
    \setlength{\tabcolsep}{3pt}
    \renewcommand{\arraystretch}{1.08}
    \begin{tabular}{ccc|c}
        \toprule
        \multicolumn{3}{c|}{Tool Choice}
        & \multirow{2}{*}{Acc. (\%)} \\
        \cmidrule(lr){1-3}
        Global Scan
        & Segment Focus
        & Temporal Stitch
        & \\
        \midrule
        \(\times\)
        & \(\checkmark\)
        & \(\checkmark\)
        & 93.23 \\
        \(\checkmark\)
        & \(\times\)
        & \(\checkmark\)
        & 92.43 \\
        \(\checkmark\)
        & \(\checkmark\)
        & \(\times\)
        & 93.63 \\
        \midrule
        \(\checkmark\)
        & \(\checkmark\)
        & \(\checkmark\)
        & \textbf{94.42} \\
        \bottomrule
    \end{tabular}
    \caption{
        Ablation study of the video observation tools on the UCF-Crime
        subset of VAU-Bench.
    }
    \label{tab:ablation_tools}
\end{table}

\begin{table}[t]
    \centering
    \small
    \setlength{\tabcolsep}{7pt}
    \renewcommand{\arraystretch}{1.05}
    \begin{tabular}{l l c}
        \toprule
        Reasoning Model & Video Model & Acc. (\%) \\
        \midrule

        \multirow{3}{*}{DeepSeek-V4-Flash}
        & Qwen2.5-VL-3B & 93.63 \\
        & Qwen2.5-VL-7B & 93.63 \\
        & Qwen3.5-4B    & \textbf{94.82} \\

        \midrule

        \multirow{3}{*}{DeepSeek-V4-Pro}
        & Qwen2.5-VL-3B & 94.42 \\
        & Qwen2.5-VL-7B & 94.82 \\
        & Qwen3.5-4B    & \textbf{95.62} \\

        \bottomrule
    \end{tabular}
    \caption{
        Ablation study with different reasoning and video models on the UCF-Crime subset of VAU-Bench.
    }
    \label{tab:ablation_models}
\end{table}

\paragraph{Temporal Anomaly Grounding.}
As reported in Table~\ref{tab:main_grounding}, AgenticVAU consistently outperforms the zero-shot baseline in temporal anomaly grounding across all three datasets.
The improvement is particularly clear on ECVA and MSAD, where both mIoU and recall at stricter overlap thresholds increase substantially.
The results demonstrate that broad exploration followed by denser inspection of candidate intervals enables the framework to recover more precise anomaly boundaries rather than merely identifying approximate anomalous regions.

\subsection{Insightful Analysis}

\subsubsection{Ablation of observation tools.}
As shown in Table~\ref{tab:ablation_tools}, removing any observation tool degrades performance, demonstrating their complementary roles.
Segment Focus contributes the most, with its removal reducing accuracy from 94.42\% to 92.43\%, highlighting the importance of dense local verification.
Global Scan and Temporal Stitch also improve performance by supporting broad exploration and cross-interval comparison, respectively.

\begin{figure*}[t]
\centering
\includegraphics[width=\textwidth]{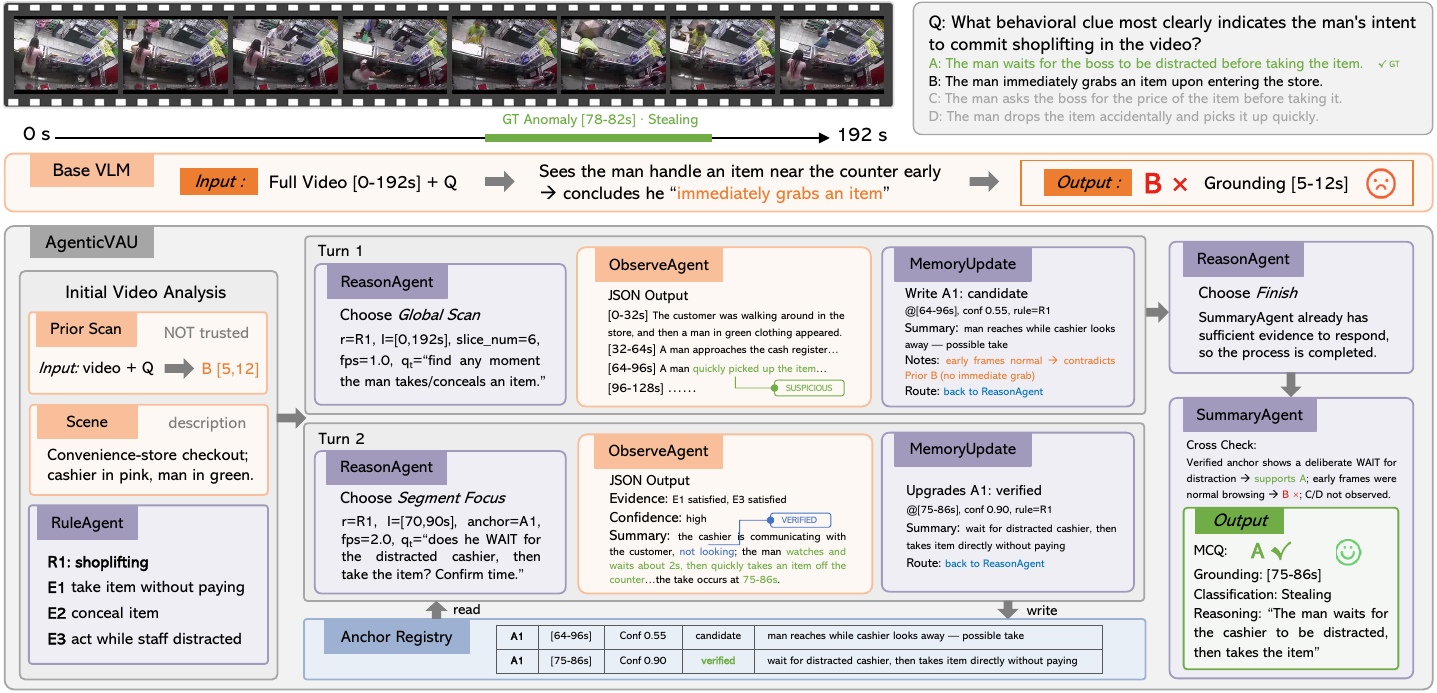}
\caption{
\textbf{Case study of AgenticVAU on a shoplifting video.}
The base VLM mistakes early normal behavior for the target event, whereas AgenticVAU uses global exploration and local verification to revise the initial prior.
The resulting verified anchor supports the correct answer, temporal grounding, anomaly class, and evidence-based explanation.
}
\label{fig:case}
\end{figure*}

\subsubsection{Ablation of model variants.}
As shown in Table~\ref{tab:ablation_models}, AgenticVAU achieves strong performance across different reasoning and video models.
Stronger backbones generally bring further improvements, with DeepSeek-V4-Pro and Qwen3.5-4B achieving the best accuracy of 95.62\%.
This demonstrates the robustness of the framework to model choices.

\subsubsection{Error Analysis.}

\begin{figure}[t]
\centering
\includegraphics[width=0.98\linewidth]{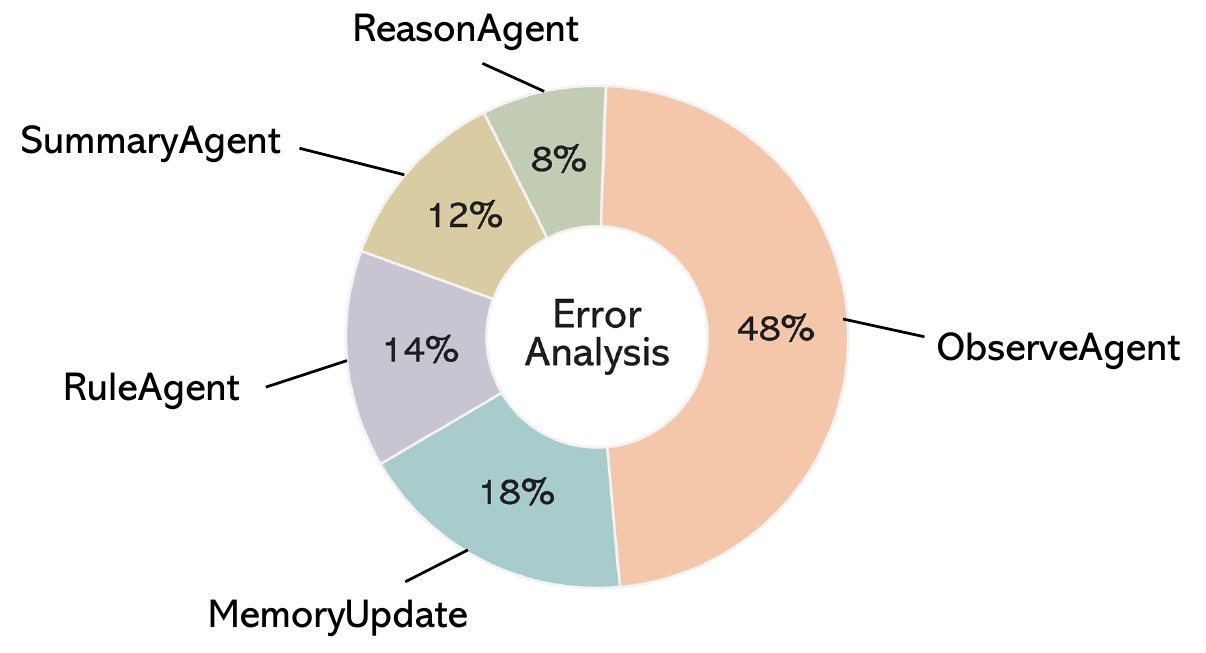}
\caption{
Distribution of the primary failure sources among 50 randomly sampled error cases.
}
\label{fig:error}
\end{figure}

To analyze the failure modes of AgenticVAU, we manually categorized 50 sampled errors by their primary source.
As shown in Figure~\ref{fig:error}, ObserveAgent accounts for most failures (48\%), followed by MemoryUpdate (18\%), RuleAgent (14\%), SummaryAgent (12\%), and ReasonAgent (8\%).
This suggests that the main bottlenecks lie in visual evidence extraction and reliable registry updates.

\subsection{Case study}

Figure~\ref{fig:case} presents a shoplifting case with a ground-truth anomaly interval of 78--82\,s.
The base VLM relies on an early observation and incorrectly selects option B with a grounding result of 5--12\,s.
In contrast, AgenticVAU treats this initial prediction as an untrusted prior and performs a global scan to identify a candidate interval.
It then applies Segment Focus to verify that the man waits until the cashier is distracted before taking the item without paying.
Accordingly, the anchor registry updates the evidence from candidate to verified, enabling the system to select the correct option A, classify the event as stealing, and localize it to 75--86\,s.
For clarity, we provide a more detailed explanation of this case in the supplementary material.

\section{Conclusion}
We presented AgenticVAU, a training-free multi-agent framework that formulates video anomaly understanding as iterative evidence exploration and verification.
Rather than relying on a fixed set of video observations, AgenticVAU progressively discovers candidate events and examines them through targeted observations.
By separating rule construction, search planning, video observation, and final decision-making, and coordinating them through the anchor registry, the framework actively collects and organizes evidence before making predictions.
Experiments on VAU-Bench show improvements over zero-shot inference and reinforcement-fine-tuned baselines across multiple VAU tasks, with particularly strong gains in anomaly classification.
These results highlight the importance of multi-agent collaboration for reliable video anomaly understanding.
Future work will focus on reducing inference cost and improving the reliability of visual observations and evidence-state updates.

\bibliography{aaai}


\end{document}